\documentclass[runningheads]{llncs}
\usepackage{graphicx}

\usepackage{subfigure}
\usepackage{bbm}
\usepackage{amsmath}
\usepackage{amssymb}
\usepackage{bbding}
\usepackage{colortbl}
\usepackage[
    colorlinks,
    linkcolor=red,
    citecolor=blue,
    urlcolor=blue
]{hyperref}
\usepackage{booktabs}
\usepackage{multirow}
\usepackage[misc]{ifsym}

\def\eg{\textit{e.g.}}

\begin{document}
\title{Foundation Model for Endoscopy Video Analysis via Large-scale Self-supervised Pre-train}
\titlerunning{Foundation Model for Endoscopy Video Analysis}
%

\author{
Zhao Wang\inst{1}$^*$
\and
Chang Liu\inst{2}$^*$
\and
Shaoting Zhang\inst{3}$^\dagger$\textsuperscript{(\Letter)}
\and
Qi Dou\inst{1}$^\dagger$\textsuperscript{(\Letter)}
}
\authorrunning{Z. Wang et al.}
\institute{
The Chinese University of Hong Kong, Hong Kong, China\\
\and
SenseTime Research, Shanghai, China\\
\and
Shanghai Artificial Intelligence Laboratory, Shanghai, China\\
}
\def\thefootnote{$*$}\footnotetext{Equal contributions.}
\def\thefootnote{$\dagger$}\footnotetext{Corresponding authors.}

%
\maketitle              

\begin{abstract}
Foundation models have exhibited remarkable success in various applications, such as disease diagnosis and text report generation. To date, a foundation model for endoscopic video analysis is still lacking. In this paper, we propose Endo-FM, a foundation model specifically developed using massive endoscopic video data. First, we build a video transformer, which captures both local and global long-range dependencies across spatial and temporal dimensions. Second, we pre-train our transformer model using global and local views via a self-supervised manner, aiming to make it robust to spatial-temporal variations and discriminative across different scenes. To develop the foundation model, we construct a large-scale endoscopy video dataset by combining 9 publicly available datasets and a privately collected dataset from Baoshan Branch of Renji Hospital in Shanghai, China. Our dataset overall consists of over 33K video clips with up to 5 million frames, encompassing various protocols, target organs, and disease types. Our pre-trained Endo-FM can be easily adopted for a given downstream task via fine-tuning by serving as the backbone. With experiments on 3 different types of downstream tasks, including classification, segmentation, and detection, our Endo-FM surpasses the current state-of-the-art (SOTA) self-supervised pre-training and adapter-based transfer learning methods by a significant margin, such as VCL (3.1\% F1, 4.8\% Dice, and 5.5\% F1 for classification, segmentation, and detection) and ST-Adapter (5.9\% F1, 9.6\% Dice, and 9.9\% F1 for classification, segmentation, and detection). Code, datasets, and models are released at \url{https://github.com/med-air/Endo-FM}.

\keywords{Foundation model  \and Endoscopy video \and Pre-train.}
\end{abstract}
%
%
%

\section{Introduction}

Foundation models pre-trained on large-scale data have recently showed success in various downstream tasks on medical images including classification \cite{fu2023anatomy}, detection \cite{willemink2022toward}, and segmentation \cite{tang2022self}.
However, medical data have various imaging modalities, and clinical data collection is expensive. It is arguable that a specific foundation model trained on some certain type of data is useful at the moment.
In this paper, we focus on endoscopic video, which is a routine imaging modality and increasingly studied in gastrointestinal disease diagnosis, minimally invasive procedure and robotic surgery. Having an effective foundation model is promising to facilitate downstream tasks that necessitate endoscopic video analysis.

Existing work on foundation models for medical tasks, such as X-ray diagnosis \cite{boecking2022making} and radiology report generation \cite{moon2022multi,moor2023foundation}, involves pre-training on large-scale image-text pairs and relies on large language models to learn cross-modality features.
However, since clinical routines for endoscopy videos typically do not involve text data, a pure image-based foundation model is currently more feasible.
To this end, we develop a video transformer, based on ViT B/16 \cite{dosovitskiy2021an}, containing 121M parameters, which serves as the foundation model backbone for our video data.
We note that a similarly scaled foundation model in recent work \cite{willemink2022toward} based on Swin UNETR \cite{hatamizadeh2022swin} with 62M parameters has been successfully employed for CT scans. This would indicate that our video transformer could have sufficient capacity to model the rich spatial-temporal information of endoscopy videos.

To learn rich spatial-temporal information from endoscopy video data \cite{hu2021contrast}, our Endo-FM is pre-trained via a self-supervised manner by narrowing the gap between feature representations from different spatial-temporal views of the same video.
These views are generated to address the variety of context information and motions of endoscopy videos.
Drawing inspiration from self-supervised vision transformers \cite{caron2021emerging,ranasinghe2022selfsupervised}, we propose to pre-train the model via a teacher-student scheme. Under this scheme, the student is trained to predict (match) the teacher's output in the latent feature space. 
In other words, given two spatial-temporal aware views from the same video, one view processed by the teacher is predicted by another one processed by the student to learn the spatial-temporal information. 
Therefore, designing effective and suitable matching strategies for different spatial-temporal views from the same endoscopy video is important.

In this paper, we propose Endo-FM, a novel foundation model designed for endoscopic video analysis. First, we build a video transformer based on ViT \cite{dosovitskiy2021an} to capture long-range spatial and temporal dependencies, together with dynamic spatial-temporal positional encoding designed for tackling input data with diverse spatial sizes and temporal frame rates. Second, Endo-FM is pre-trained under a teacher-student scheme via spatial-temporal matching on diverse video views. Specifically, we create various spatial-temporal aware views differing in spatial sizes and frame rates for an input video clip. Both teacher and student models process these views of a video and predict one view from another in the latent feature space. This enables Endo-FM to learn \emph{spatial-temporal invariant} (to view, scale, and motion) features that are transferable across different endoscopic domains and disease types while retaining discriminative features that are specific to each context. We construct a large-scale endoscopic video dataset by combining 9 public and a new private collected dataset from Baoshan Branch of Renji Hospital in Shanghai, China, with over 33K video clips with up to 5 million frames. Our pre-trained Endo-FM can be easily applied to various downstream tasks by serving as the backbone. Experimental results on 3 different types of downstream tasks demonstrate the effectiveness of Endo-FM, surpassing the current state-of-the-art self-supervised pre-training and adapter-based transfer learning methods by a significant margin, such as VCL (3.1\% F1, 4.8\% Dice, and 5.5\% F1 for classification, segmentation, and detection) and ST-Adapter (5.9\% F1, 9.6\% Dice, and 9.9\% F1 for classification, segmentation, and detection).

\section{Method}

\begin{figure}[t]
\centering
\includegraphics[width=\textwidth]{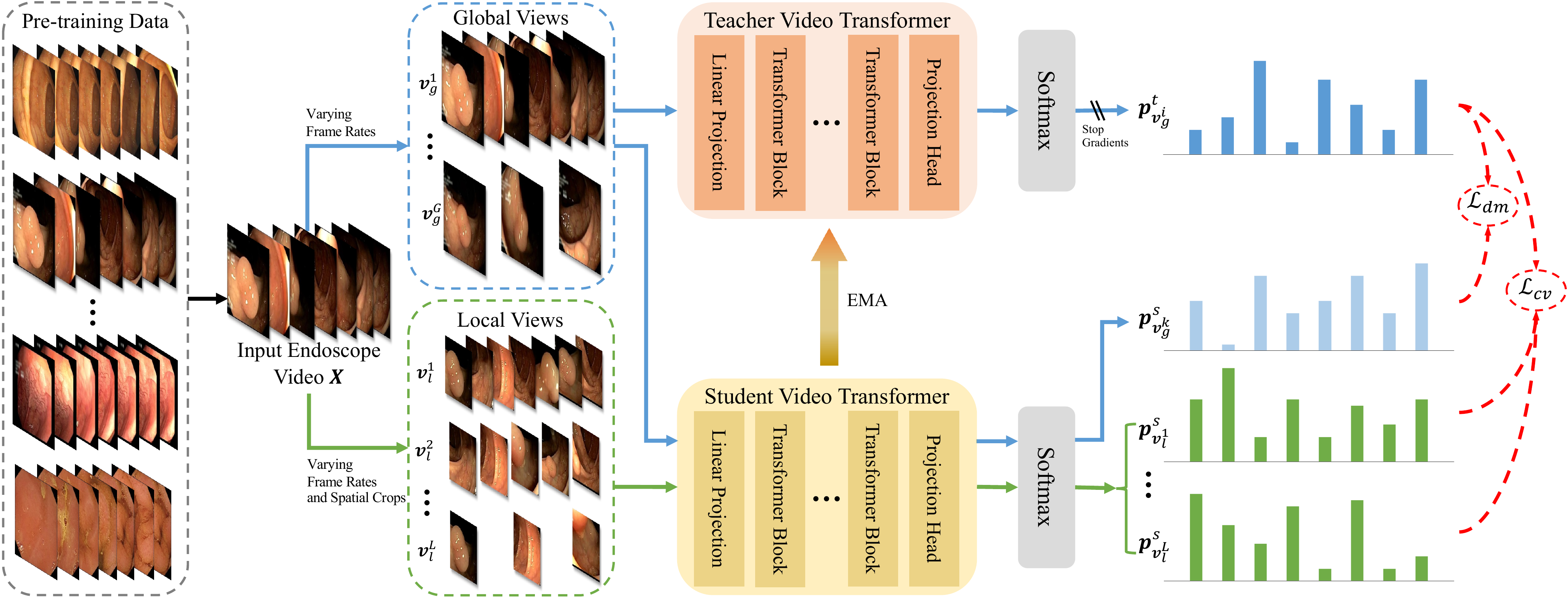}
\caption{
Illustration of our proposed Endo-FM. We build a video transformer model and design a self-supervised pre-training approach.
} 
\label{fig:framework}
\end{figure}

To begin with, we build a video transformer as the architecture of our Endo-FM (Sec. \ref{subsection:video}). Then, we propose a novel self-supervised spatial-temporal matching scheme (Sec. \ref{subsection:pretrain}). Finally, we describe the overall training objective and specifics in Sec. \ref{subsection:objective}. An overview of our method is shown in Figure \ref{fig:framework}.

\subsection{Video Transformer for Spatial-Temporal Encoding}
\label{subsection:video}

We build a video transformer to encode input endoscopic video. 
The spatial and temporal attention mechanisms in our model capture long-range dependencies across both spatial and temporal dimensions, with a larger receptive field than conventional convolutional kernels \cite{naseer2021intriguing}.
Our model is built using 12 encoder blocks, equipped with space-time attention \cite{bertasius2021space}.
Specifically, given an endoscopic video clip $\boldsymbol{X}\!\in\!\mathbb{R}^{T \!\times\! 3 \!\times\! H \!\times\! W}$ as input, consisting of $T$ frames with size $H\!\!\times\!\!W$, each frame in $\boldsymbol{X}$ is divided into $N\!=\!HW\!/\!P^2$ patches of size $P\!\!\times\!\!P$, and these patches are then mapped into $N$ patch tokens. 
Thus, each encoder block processes $N$ patch (spatial) and $T$ temporal tokens. 
Given the intermediate token $z^m\!\in\!\mathbb{R}^{D}$ for a patch from block $m$, the token computation in the next block is as follows:
\begin{equation}
\small
\begin{aligned}
\boldsymbol{z}_{\operatorname{time}}^{m+1} & =\operatorname{MHSA}_{\operatorname{time}}\left(\operatorname{L N}\left(\boldsymbol{z}^m\right)\right)+\boldsymbol{z}^m, \\
\boldsymbol{z}_{\operatorname{space}}^{m+1} & =\operatorname{MHSA}_{\operatorname{space}}\left(\operatorname{LN}\left(\boldsymbol{z}_{\operatorname{time}}^{m+1}\right)\right)+\boldsymbol{z}_{\operatorname{time}}^{m+1}, \\
\boldsymbol{z}^{m+1} & =\operatorname{MLP}\left(\operatorname{LN}\left(\boldsymbol{z}_{\operatorname{space}}^{m+1}\right)\right)+\boldsymbol{z}_{\operatorname{space}}^{m+1},
\end{aligned}
\label{equ:spatial_temporal_encoding}
\end{equation}
where MHSA denotes multi-head self-attention, LN denotes LayerNorm \cite{ba2016layer}, and MLP denotes multi-layer perceptron.
Our model also includes a learnable class token, representing the global features learned by the model along the spatial and temporal dimensions. 
For pre-training, we use a MLP to project the class token from the last encoder block as the feature $\boldsymbol{f}$ of $\boldsymbol{X}$.

Different from static positional encoding in ViT \cite{dosovitskiy2021an}, we design a dynamic spatial-temporal encoding strategy to help our model tackle various spatial-temporal views with different spatial sizes and frame rates (Sec. \ref{subsection:pretrain}).
Specifically, We fix the spatial and temporal positional encoding vectors to the highest resolution of the input view for each dimension, making it easy to interpolate for views with smaller spatial size or lower temporal frame rate.
These spatial and temporal positional encoding vectors are added to the corresponding spatial and temporal tokens.
Such dynamic strategy ensures that the learned positional encoding is suitable for downstream tasks with diverse input sizes.

\subsection{Self-supervised Pre-train via Spatial-Temporal Matching}

\label{subsection:pretrain}
Considering the difficulties of tackling the context information related with lesions, tissues, and dynamic scenes in endoscopic data, we pre-train Endo-FM to be robust to such spatial-temporal characteristics.
Inspired by self-supervised vision transformers \cite{caron2021emerging}, the pre-training is designed in a teacher-student scheme, where the student is trained to match the teacher’s output.
To achieve this, given an input video $\boldsymbol{X}$, we create two types of spatial-temporal views serving as the model inputs: global and local views, as shown in Fig. \ref{fig:framework}.
The global views $\small\{\boldsymbol{v}_g^i\!\!\in\!\!\mathbb{R}^{T_{g}^i \times 3 \times H_g \times W_g}\}_{i=1}^{G}$ are generated by uniformly sampling $\boldsymbol{X}$ with different frame rates, and the local ones $\small\{\boldsymbol{v}_l^j\!\!\in\!\!\mathbb{R}^{T_{l}^j \times 3 \times H_l \times W_l}\}_{j=1}^{L}$ are generated by uniformly sampling video frames with different frame rates from a randomly cropped region of $\boldsymbol{X}$ ($\small T_l \leq T_g$). 
During pre-training, the global views are fed into both teacher and student, and the local ones are only fed into the student.
The model output $\boldsymbol{f}$ is then normalized by a softmax function with a temperature $\tau$ to obtain the probability distribution $\small\boldsymbol{p}\!\!=\!\!\operatorname{softmax}(\boldsymbol{f}/\tau)$.
In the following, we design two matching schemes with respect to the difficulties of tackling endoscopy videos.

\noindent {\bf Cross-view Matching.}
Different from image-based pre-training \cite{willemink2022toward}, our video-oriented pre-training is designed to capture the relationships between different spatial-temporal variations. Specifically, the context information presented in different frames of the same endoscope video can vary under two key factors: 1) the proportion of tissue and lesions within the frame, and 2) the presence or absence of lesion areas.
To address these, we employ a \emph{cross-view matching} approach where the target global views processed by the teacher ($\small\{\boldsymbol{p}^t_{\boldsymbol{v}_g^i}\}_{i=1}^{G}$) are predicted from the online local views processed by the student ($\small\{\boldsymbol{p}^s_{\boldsymbol{v}_l^j}\}_{j=1}^{L}$). 
By adopting this strategy, our model learns high-level context information from two perspectives: 1) spatial context in terms of the possible neighboring tissue and lesions within a local spatial crop, and 2) temporal context in terms of the possible presence of lesions in the previous or future frames of a local temporal crop. 
Thus, our method effectively addresses the proportion and existence issues that may be encountered.
We minimize the following loss for cross-view matching:
\begin{equation}
\small
\mathcal{L}_{\operatorname{cv }}=\sum_{i=1}^G \sum_{j=1}^L - \boldsymbol{p}_{\boldsymbol{v}_g^i}^t \cdot \log \boldsymbol{p}_{\boldsymbol{v}_l^j}^s.
\label{equ:cross_view_loss}
\end{equation}

\noindent {\bf Dynamic Motion Matching.}
In addition to the proportion and existence issues of lesions, a further challenge arises from the inherent dynamic nature of the scenes captured in endoscopy videos. The speeds and ranges of motion can vary greatly across different videos, making it difficult to train a model that is effective across a wide range of dynamic scenarios.
The previous model \cite{qian2021spatiotemporal} learned from clips with fixed frame rate can not tackle this issue, as clips sampled with various frame rates contain different motion context information (\eg, fast v.s. slow scene changing) and differ in nuanced tissue and lesions.
To address this challenge, our approach involves motion modeling during pre-training under dynamic endoscope scenes by predicting a target global view ($\small\boldsymbol{p}^t_{\boldsymbol{v}_g^i}$) processed by the teacher from another online global view ($\small\boldsymbol{p}^s_{\boldsymbol{v}_g^k}$) processed by the student. Moreover, by predicting the nuanced differences of tissue and lesions in a view with a high frame rate from another with a low frame rate, the model is encouraged to learn more comprehensive motion-related contextual information.
The dynamic motion difference among global view pairs is minimized by
\begin{equation}
\small
\mathcal{L}_{\operatorname{dm}}=\sum_{i=1}^G \sum_{k=1}^G-\mathbbm{1}_{[i\neq k]}\boldsymbol{p}_{\boldsymbol{v}_g^i}^t \cdot \log \boldsymbol{p}_{\boldsymbol{v}_g^k}^s,
\label{equ:dynamic_motion_loss}
\end{equation}
where $\mathbbm{1}[\cdot]$ is an indicator function.

\subsection{Overall Optimization Objective and Pre-training Specifics}
\label{subsection:objective}
The overall training objective for Endo-FM is $\mathcal{L}_{\operatorname{pre-train}} = \mathcal{L}_{\operatorname{cv}} + \mathcal{L}_{\operatorname{dm}}$.
Centering and sharpening schemes \cite{caron2021emerging} are incorporated to the teacher outputs.
To prevent the problem of the teacher and student models constantly outputting the same value during pre-training, we update the student model $\theta$ through backpropagation, while the teacher model $\phi$ is updated through exponential moving average (EMA) using the student's weights. 
This is achieved by updating the teacher's weights as $\phi_{t} \leftarrow \alpha \phi_{t-1}+(1-\alpha) \theta_t$ at each training iteration $t$. 
Here, $\alpha$ is a momentum hyper-parameter that determines the updating rate.

Except for the challenges posed by the issues of size proportion, existence, and dynamic scenes in Sec. \ref{subsection:pretrain}, we have also observed that the appearance of endoscope videos is highly diverse. These videos are captured using different surgical systems and in a wide range of environmental conditions \cite{goodman2021real}. To address this variability, we apply temporally consistent spatial augmentations \cite{qian2021spatiotemporal} to all frames within a single view.
Our augmentation approach includes random horizontal flips, color jitter, Gaussian blur, solarization, and so on, which enhances the robustness and generalizability of Endo-FM.

For Endo-FM, we set the patch size $P$ as $16$ and embedding dimension $D$ as $768$.
We create $G\!=\!2$ global views and $L\!=\!8$ local views for every input endoscopy video, where $T_g\!\in\![8, 16]$ and $T_l\!\in\![2, 4, 8, 16]$.
The spatial sizes of global and local views are $224\!\times\!224$ and $96\!\times\!96$, respectively.
The MLP head projects the dimension of class token to $65536$.
The temperature hyper-parameters are set as $\tau_t\!=\!0.04$ and $\tau_s\!=\!0.07$.
The EMA update momentum $\alpha$ is $0.996$.
The training batch size is $12$ with AdamW \cite{loshchilov2018decoupled} optimizer (learning rate $2e\!-\!5$, weight decay $4e\!-\!2$).
The pre-training is finished with $30$ epochs with a cosine schedule \cite{loshchilov2017sgdr}.

\begin{table}[t]
\setlength\tabcolsep{3pt}
  \centering
  \caption{Details of all pre-train and downstream datasets used in this work.}
  \scalebox{0.63}{

\begin{tabular}{ll|l|l|l|l|l}
\toprule
\textbf{Phase} & \textbf{Dataset} & \textbf{Provider} & \textbf{Videos} & \textbf{Frames} & \textbf{Protocol} & \textbf{Disease} \\
\midrule
\multirow{9}[6]{*}{Pre-train} & Colonoscopic \cite{mesejo2016computer} & CNRS  & 210   & 36534 & colonoscope & adenoma, hyperplasia \\
      & SUN \cite{misawa2021development} \& SUN-SEG \cite{ji2022video} & ANU   & 1018  & 159400 & colonoscope & SSL, adenoma, hyperplasia, T1b \\
      & LDPolypVideo \cite{ma2021ldpolypvideo} & USTC  & 237   & 40186 & colonoscope & polyp \\
      & Hyper-Kvasir \cite{borgli2020hyperkvasir} & Simula & 5704  & 875940 & gastroscope & barrett's oesophagus, polyp, cancer \\
      & Kvasir-Capsule \cite{smedsrud2021kvasir} & Simula & 1000  & 158892 & gastroscope & erosion, erythema, etc. \\
      & CholecTriplet \cite{nwoye2022rendezvous} & BIDMC & 580   & 90444 & laparoscope & cholecystectomy \\
\cmidrule{2-7}      & \multirow{2}[2]{*}{Ours} & Baoshan Branch  & 16494 & 2491952 & colonoscope & \multirow{2}[2]{*}{polyp, erosion, etc.} \\
      &       & of Renji Hospital & 7653  & 1170753 & gastroscope &  \\
\cmidrule{2-7}      & Summary & 6 providers & 32896 & 5024101 & 3 protocols & 10+ diseases \\
\midrule
\multirow{4}[4]{*}{Downstream} & PolypDiag \cite{tian2022contrastive} & Adelaide & 253   & 485561 & gastroscope & polyp, cancer \\
      & CVC-12k \cite{bernal2015wm} & UAB   & 29    & 612   & colonoscope & polyp \\
      & KUMC \cite{li2021colonoscopy} & Kansas  & 53    & 19832 & colonoscope & adenoma, hyperplasia \\
\cmidrule{2-7}      & Summary & 3 providers & 335   & 506005 & 2 protocols & 4 diseases \\
\bottomrule
\end{tabular}%

}

  \label{tab:dataset}%
\end{table}%

\begin{figure}[t]
\centering
\includegraphics[width=0.95\textwidth]{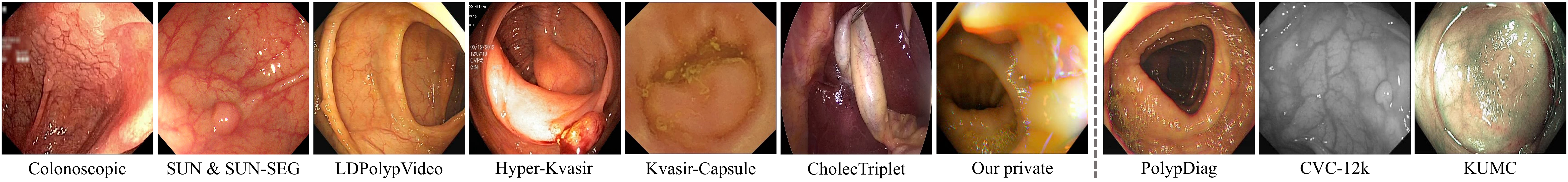}
\caption{Example frames of the 10 pre-train and downstream datasets used in this work.
} 
\label{fig:pretrain_data}
\end{figure}

\section{Experiment}

\subsection{Datasets and Downstream Setup}

We collect all possible public endoscope video datasets and a new one from Baoshan Branch of Renji Hospital for pre-training.
As shown in Table \ref{tab:dataset}, these public datasets are provided by world-wide research groups \cite{borgli2020hyperkvasir,misawa2021development,ji2022video,ma2021ldpolypvideo,mesejo2016computer,smedsrud2021kvasir} and previous EndoVis challenge \cite{nwoye2022rendezvous}, covering 3 endoscopy protocols and 10+ types of diseases.
We process the original videos into 30fps short clips with a duration of 5s on average.
We evaluate our pre-trained Endo-FM on three downstream tasks: disease diagnosis (PolypDiag \cite{tian2022contrastive}), polyp segmentation (CVC-12k \cite{bernal2015wm}), and detection (KUMC \cite{li2021colonoscopy}). 
The detailed information of three downstream datasets is shown in Table \ref{tab:dataset}.
The example frames of the 10 datasets are shown in Fig. \ref{fig:pretrain_data}.

For downstream fine-tuning, we utilize the following setup:
1) PolypDiag: A randomly initialized linear layer is appended to our pre-trained Endo-FM. 
We sample $8$ frames with spatial size $224\!\times\!224$ for every video as the input and train for $20$ epochs. 
2) CVC-12k: A TransUNet equipped with Endo-FM as the backbone is implemented.
We resize the spatial size as $224\!\times\!224$ and train for $150$ epochs. 
3) KUMC: we implement a STFT \cite{wu2021multi} with our pre-trained model as backbone for generating feature pyramid.
We resize the spatial size as $640\!\times\!640$ and train for $24$k iterations. 
We report F1 score for PolypDiag, Dice for CVC-12k, and F1 score for KUMC.

\begin{table}[t]
\setlength\tabcolsep{3pt}
  \centering
  \caption{Comparison with other latest SOTA methods on $3$ downstream tasks. 
  We report F1 score (\%) for PolypDiag, Dice (\%) for CVC-12k, and F1 score (\%) for KUMC. }

\scalebox{0.85}{

\begin{tabular}{l|c|c|ccc}
\toprule
\multicolumn{1}{c|}{\multirow{2}[2]{*}{Method}} & \multirow{2}[2]{*}{Venue} & Pre-training & PolypDiag & CVC-12k & KUMC \\
      &       & Time (h) & (Classification) & (Segmentation) & (Detection) \\
\midrule
Scratch (Rand. init.) &       & N/A   & 83.5$\pm$1.3 & 53.2$\pm$3.2 & 73.5$\pm$4.3 \\
\midrule
TimeSformer \cite{bertasius2021space} & ICML'21 & 104.0  & 84.2$\pm$0.8 & 56.3$\pm$1.5 & 75.8$\pm$2.1 \\
CORP \cite{hu2021contrast} & ICCV'21 & 65.4  & 87.1$\pm$0.6 & 68.4$\pm$1.1 & 78.2$\pm$1.4 \\
FAME \cite{ding2022motion} & CVPR'22 & 48.9  & 85.4$\pm$0.8 & 67.2$\pm$1.3 & 76.9$\pm$1.2 \\
ProViCo \cite{park2022probabilistic} & CVPR'22 & 71.2  & 86.9$\pm$0.5 & 69.0$\pm$1.5 & 78.6$\pm$1.7 \\
VCL \cite{qian2022static} & ECCV'22 & 74.9  & 87.6$\pm$0.6 & 69.1$\pm$1.2 & 78.1$\pm$1.9 \\
ST-Adapter \cite{pan2022st} & NeurIPS'22 & 8.1   & 84.8$\pm$0.7 & 64.3$\pm$1.9 & 74.9$\pm$2.9 \\
\midrule
\rowcolor[rgb]{ .914,  .914,  .914} \textbf{Endo-FM (Ours)} &       & 20.4  & \boldmath{}\textbf{90.7$\pm$0.4}\unboldmath{} & \boldmath{}\textbf{73.9$\pm$1.2}\unboldmath{} & \boldmath{}\textbf{84.1$\pm$1.3}\unboldmath{} \\
\bottomrule
\end{tabular}%

}

  \label{tab:sota_comp}%
\end{table}%

\subsection{Comparison with State-of-the-art Methods}
We compare our method with recent SOTA video-based pre-training methods, including the {\bf TimeSformer} \cite{bertasius2021space} introduces spatial-temporal attention for video processing, the {\bf CORP} \cite{hu2021contrast} presents a self-supervised contrast-and-order representation framework, the {\bf FAME}\cite{ding2022motion} proposes a foreground-background merging scheme, the {\bf ProViCo} \cite{park2022probabilistic} applies a self-supervised probabilistic video contrastive learning strategy, the {\bf VCL} \cite{qian2022static} learns the static and dynamic visual concepts, and the {\bf ST-Adapter} \cite{pan2022st} adapts the CLIP \cite{radford2021learning} by a depth-wise convolution.
We also train our model from scratch to serve as a baseline.
The same experimental setup is applied to all the experiments for fair comparisons.

Quantitative comparison results are shown in Table \ref{tab:sota_comp}.
We can observe that the scratch model shows low performance on all 3 downstream tasks, especially for segmentation. 
Compared with training from scratch, our Endo-FM achieves $+7.2\%$ F1, $+20.7\%$ Dice, and $+10.6\%$ F1 improvements for classification, segmentation, and detection tasks, respectively, indicating the high effectiveness of our proposed pre-training approach.
Moreover, our Endo-FM outperforms all SOTA methods, with $+3.1\%$ F1, $+4.8\%$ Dice, and $+5.5\%$ F1 boosts for the 3 downstream tasks over the second-best.
Such significant improvements are benefited from our specific spatial-temporal pre-training designed for endoscopy videos to tackle the complex context information and dynamic scenes.
Meanwhile, Endo-FM requires less pre-training time than SOTA pre-training methods, except the lighter but much worse ST-Adapter \cite{pan2022st}.

\subsection{Analytical Studies}

Without loss of generality, we conduct ablation studies on polyp diagnosis task from 3 aspects:
1) components analysis of our pre-training method;
2) varying combinations of global and local views in spatial-temporal matching;
3) varying the construction of global and local views.

\begin{figure}[t]
\centering

\subfigure[]{
\begin{minipage}[t]{0.23\textwidth}
\label{fig:components}
\centering
\includegraphics[width=\textwidth]{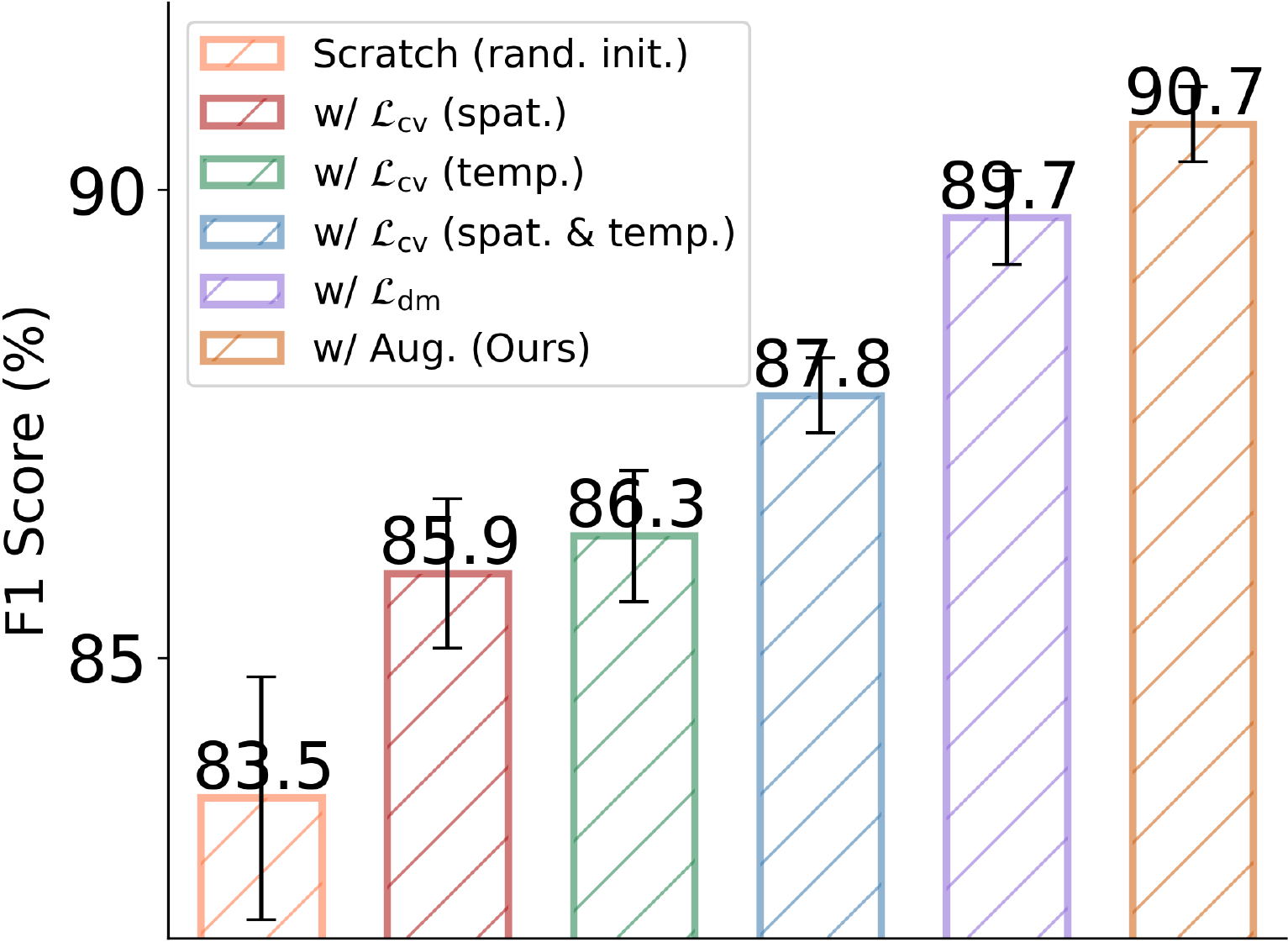}
\end{minipage}
}%
\subfigure[]{
\begin{minipage}[t]{0.23\textwidth}
\label{fig:combinations}
\centering
\includegraphics[width=\textwidth]{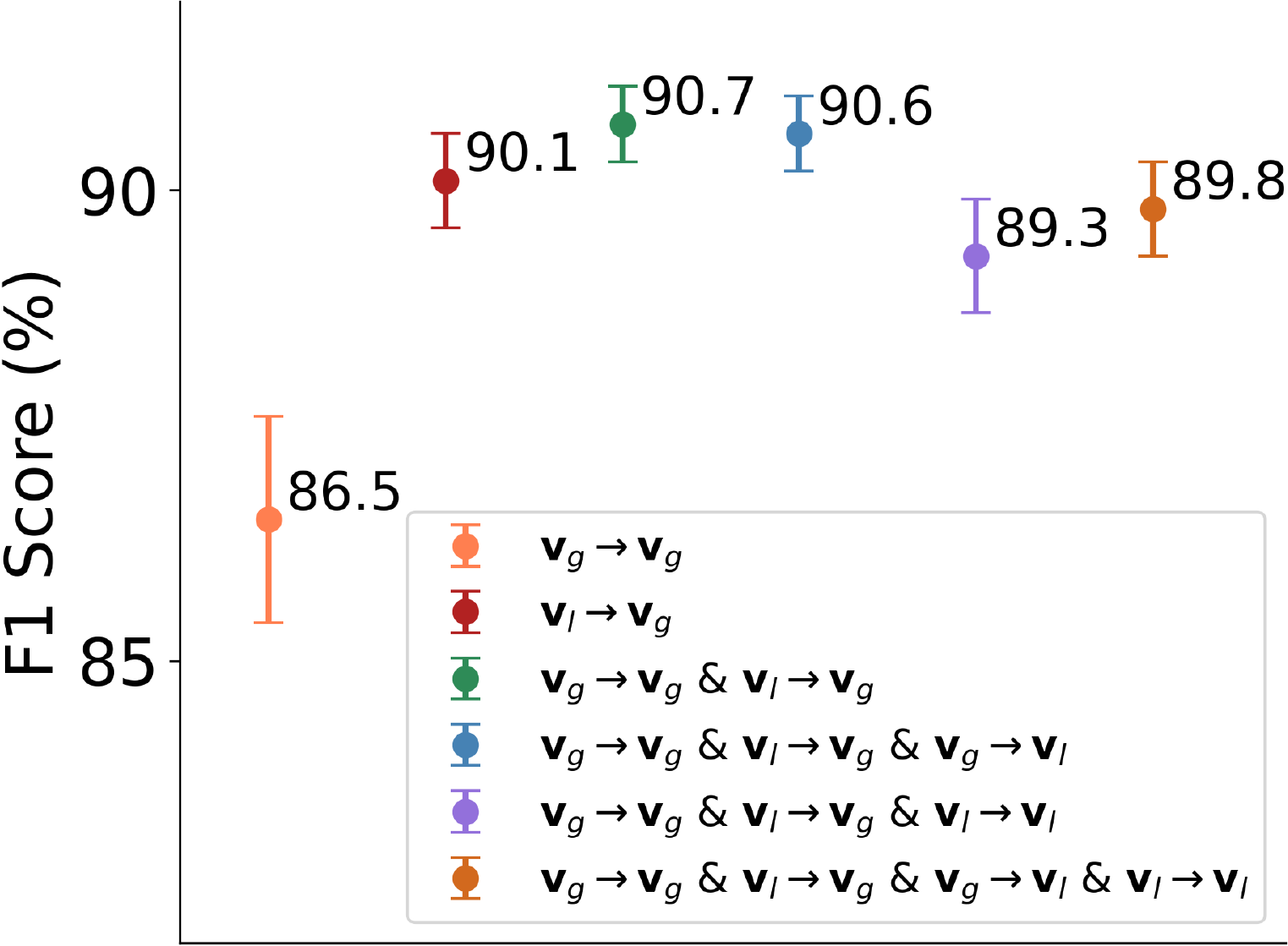}
\end{minipage}%
}%
\subfigure[]{
\begin{minipage}[t]{0.23\textwidth}
\label{fig:localviews}
\centering
\includegraphics[width=\textwidth]{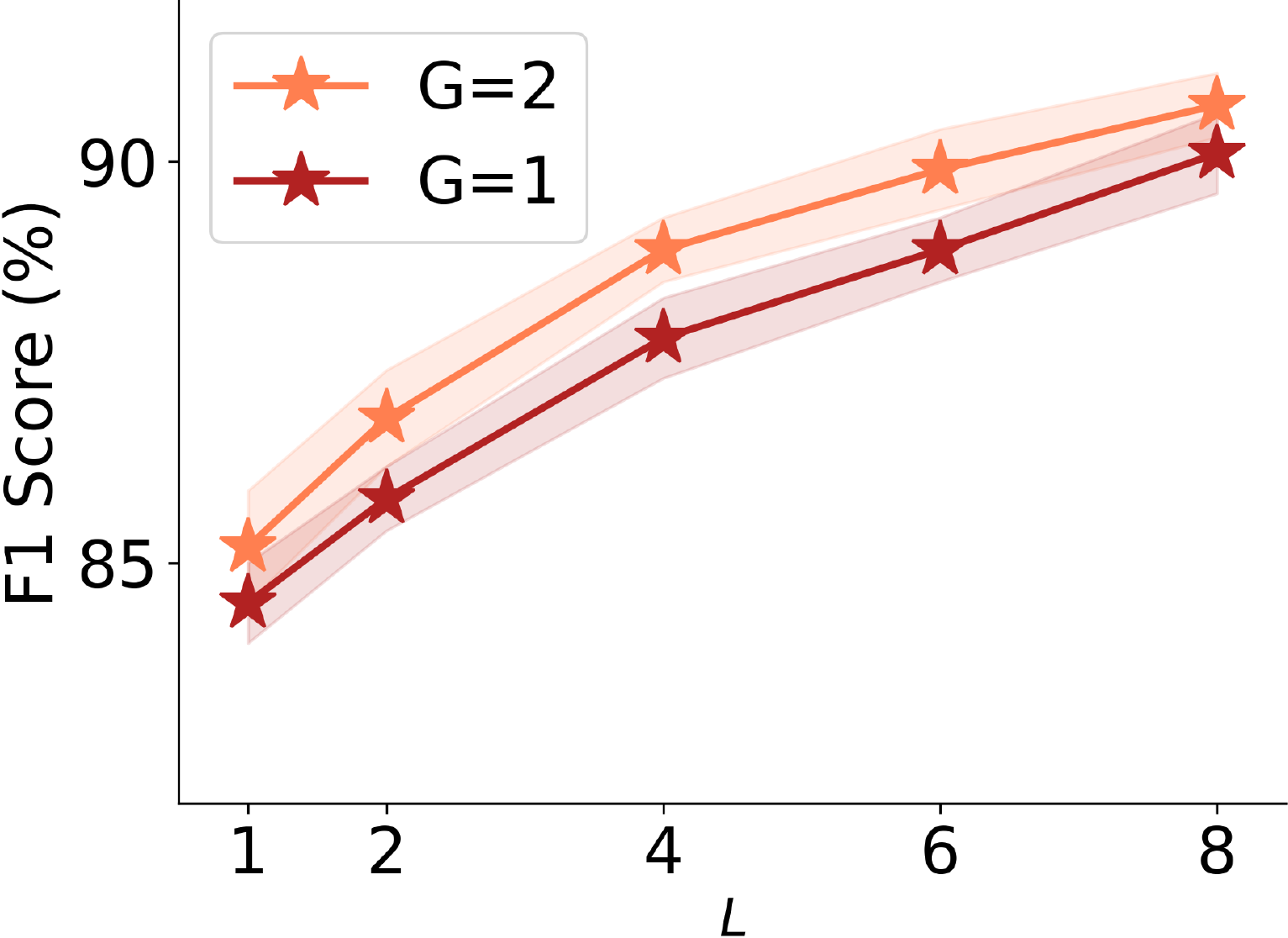}
\end{minipage}%
}%
\subfigure[]{
\begin{minipage}[t]{0.23\textwidth}
\label{fig:framenumber}
\centering
\includegraphics[width=\textwidth]{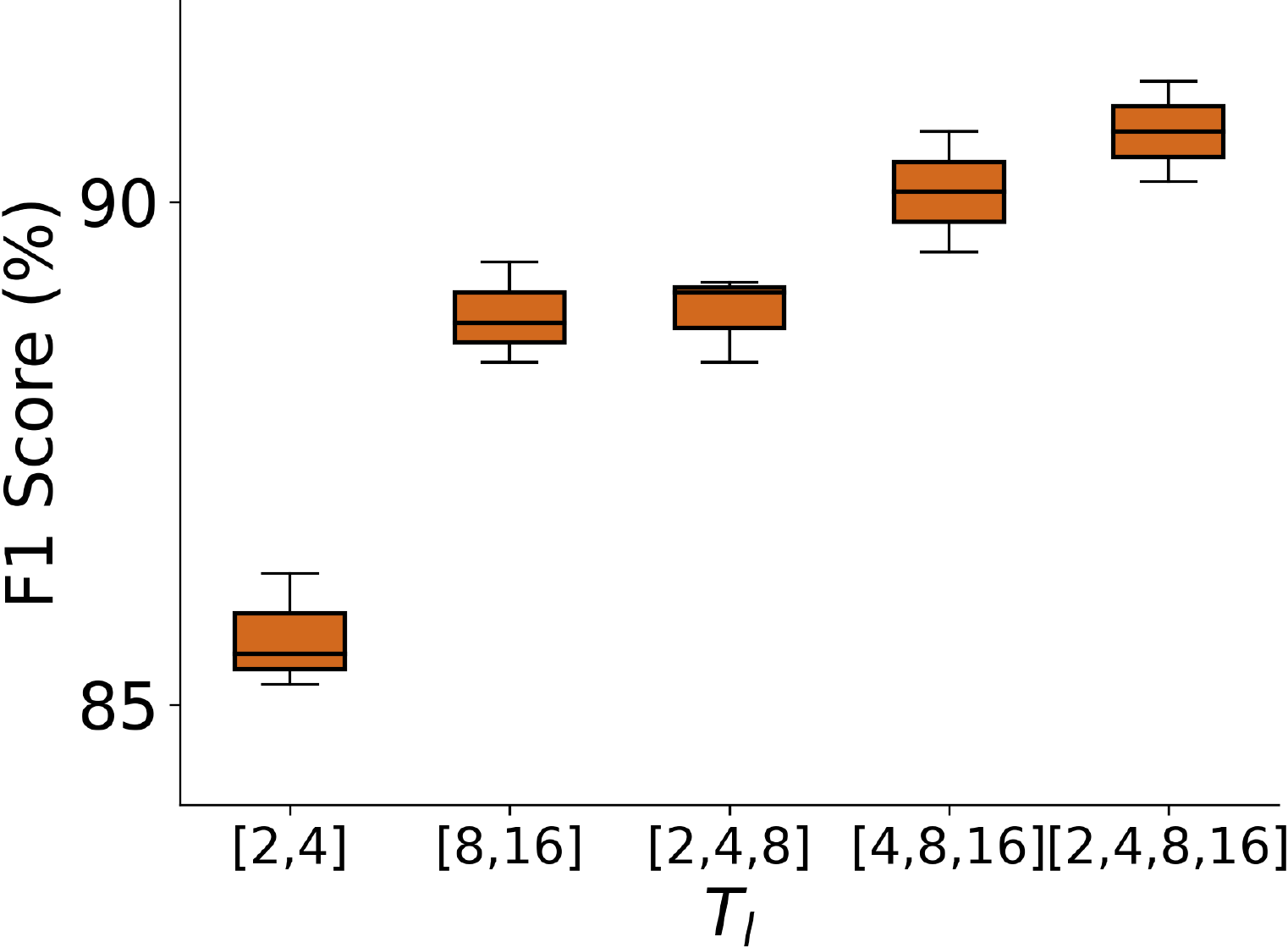}
\end{minipage}%
}%

\centering
\caption{
Ablations on PolypDiag: (a) components analysis; (b) different combinations of views; (c) number of global and local views; (d) length of local views.
}

\end{figure}

\noindent{\bf Components Analysis.}
We first study each component in our approach, as shown in Fig. \ref{fig:components}.
Here, ``w\!/\! $\mathcal{L}_{\operatorname{cv}}$ (spat.)'' and ``w\!/\! $\mathcal{L}_{\operatorname{cv}}$ (temp.)'' indicate that only spatial and temporal sampling are used for generating the local views.
We can learn that both spatial and temporal sampling for local views can help improve the performance and their combination produces a plus, yielding $+4.3\%$ F1 improvement.
Furthermore, our proposed dynamic matching scheme boosts the performance to $89.7\%$, demonstrating the importance of capturing the motion related context information from dynamic scenes.
Additionally, the performance is further improved with video augmentations from $89.7\%$ to $90.7\%$.

\noindent{\bf Spatial-temporal Matching Combinations.}
We further investigate the effects of combinations of global and local views in spatial-temporal matching, as depicted in Fig. \ref{fig:combinations}.
Here, the notation $\boldsymbol{v}_l \rightarrow \boldsymbol{v}_g$ represents the prediction of $\boldsymbol{v}_g$ from $\boldsymbol{v}_l$, and vice versa.
It indicates that joint prediction scenarios, where we predict $\boldsymbol{v}_g$ from both $\boldsymbol{v}_l$ (cross-view matching) and $\boldsymbol{v}_g$ (dynamic motion matching), result in optimal performance.
This trend can be attributed to the fact that joint prediction scenarios allow for a more comprehensive understanding of the context in complex endoscopy videos, which is lacking in individual cases.

\noindent{\bf Construction of Global and Local Views.}
We conduct a further analysis of the strategies for constructing global ($G\!\in\![1, 2]$) and local views ($L\!\in\![1, 2, 4, 6, 8]$). 
We vary the number of global and local views, and the length of local views ($T_l$), as depicted in Fig. \ref{fig:localviews}, and Fig. \ref{fig:framenumber}.
We find that incorporating more views and increasing the length variations of local views yields better performance.
For ``$G\!=\!1$'', we still create $2$ global views for $\mathcal{L}_{\operatorname{dm}}$ but only consider the longer one for $\mathcal{L}_{\operatorname{cv}}$.
These improvements stem from the spatial-temporal change invariant and cross-video discriminative features learned from the diverse endoscopy videos.

\section{Conclusion and Discussion}

To the best of our knowledge, we develop the first foundation model, Endo-FM, Which is specifically designed for analyzing endoscopy videos. 
Endo-FM is built upon a video transformer to capture rich spatial-temporal information and pre-trained to be robust to diverse spatial-temporal variations.
A large-scale endoscope video dataset with over 33K video clips is constructed.
Extensive experimental results on 3 downstream tasks demonstrate the effectiveness of Endo-FM, significantly outperforming other state-of-the-art video-based pre-training methods, and showcasing its potential for clinical application.

Regarding the recent SAM \cite{kirillov2023segany} model, which is developed for segmentation task, we try to apply SAM for our downstream task CVC-12k with the same fine-tuning scheme as Endo-FM. The experimental results show that SAM can achieve comparable performance with our Endo-FM for the downstream segmentation task. Considering that SAM is trained with 10x samples, our domain Endo-FM is considered to be powerful for endoscopy scenarios. Moreover, besides segmentation, Endo-FM can also be easily applied to other types of tasks including classification and detection. Therefore, we envision that, despite existence of general-purpose foundation models, Endo-FM or similar domain-specific foundation models will be helpful for medical applications.

\paragraph{\bf Acknowledgements.} This work was supported in part by Shenzhen Portion of Shenzhen-Hong Kong Science and Technology Innovation Cooperation Zone under HZQB-KCZYB-20200089, in part by Science, Technology and Innovation Commission of Shenzhen Municipality Project No. SGDX20220530111201008, in part by Hong Kong Innovation and Technology Commission Project No. ITS/237/21FP, in part by Hong Kong Research Grants Council Project No. T45-401/22-N, in part by the Action Plan of Shanghai Science and Technology Commission [21SQBS02300], and in part by National Key R\&D Program of China Project 2022ZD0161100.

\bibliographystyle{splncs04}
\bibliography{paper676}

\end{document}